\documentclass[letterpaper, 10 pt, conference]{ieeeconf}  % Comment this line out if you need a4paper
\usepackage{amsmath}
\usepackage{amsfonts}
\usepackage{amssymb}
\usepackage{comment}
\usepackage{xcolor}
\usepackage[ruled, vlined]{algorithm2e}
\usepackage{threeparttable}

\SetCommentSty{mycommfont}

\IEEEoverridecommandlockouts                              % This command is only needed if 
\usepackage{graphicx} % for pdf, bitmapped graphics files

\title{\LARGE \bf
Combining Reinforcement Learning with Model Predictive Control for On-Ramp Merging}

\author{Joseph Lubars$^{1}$, Harsh Gupta$^{1}$, Sandeep Chinchali$^{3}$, Liyun Li$^{2}$, Adnan Raja$^{2}$, R. Srikant$^{1}$, and Xinzhou Wu$^{2}$%
\thanks{$^{1}$Joseph Lubars ({\tt\small lubars2@illinois.edu}), Harsh Gupta ({\tt\small hgupta10@illinois.edu}) and R. Srikant ({\tt\small rsrikant@illinois.edu}) are with the Department of Electrical and Computer Engineering and the Coordinated Science Laboratory, University of Illinois at Urbana-Champaign.
        }%
\thanks{$^{2}$Adnan Raja ({\tt\small adnanr@xmotors.ai}), Liyun Li ({\tt\small liyunl@xmotors.ai}) and Xinzhou Wu ({\tt\small wuxz@xmotors.ai}) are with Xmotors.ai.}%
\thanks{$^{3}$Sandeep Chinchali ({\tt\small sandeepc@utexas.edu}) is with the department of Electrical and Computer Engineering at the University of Texas at Austin}%
}

\begin{document}

\maketitle
\thispagestyle{empty}
\pagestyle{empty}

%%%%%%%%%%%%%%%%%%%%%%%%%%%%%%%%%%%%%%%%%%%%%%%%%%%%%%%%%%%%%%%%%%%%%%%%%%%%%%%%
\begin{abstract}
We consider the problem of designing an algorithm to allow a car to autonomously merge on to a highway from an on-ramp. Two broad classes of techniques have been proposed to solve motion planning problems in autonomous driving: Model Predictive Control (MPC) and Reinforcement Learning (RL). In this paper, we first establish the strengths and weaknesses of state-of-the-art MPC and RL-based techniques through simulations. We show that the performance of the RL agent is worse than that of the MPC solution from the perspective of safety and robustness to out-of-distribution traffic patterns, i.e., traffic patterns which were not seen by the RL agent during training. On the other hand, the performance of the RL agent is better than that of the MPC solution when it comes to efficiency and passenger comfort. We subsequently present an algorithm which blends the model-free RL agent with the MPC solution and show that it provides better trade-offs between all metrics -- passenger comfort, efficiency, crash rate and robustness.
% -- compared to the standalone solutions.
\end{abstract}

%%%%%%%%%%%%%%%%%%%%%%%%%%%%%%%%%%%%%%%%%%%%%%%%%%%%%%%%%%%%%%%%%%%%%%%%%%%%%%%%
\section{INTRODUCTION}

% Page Limit: 6 pages, plus unlimited references

The domain of autonomous driving has seen rapid progress over the last few years, both in academic research as well as in industrial development. Several big companies such as Alphabet, Tesla, General Motors, and NVIDIA have poured a tremendous amount of resources towards the goal of making fully autonomous vehicles. Although a significant amount of progress has already been made, fully autonomous driving is not in the market yet, but many components of autonomous driving are already in the market to help drivers perform limited maneuvers, including automated lane changing and parking. One important problem that is yet to be fully addressed is that of merging on to a highway from an on-ramp. This problem has piqued a lot of interest recently, owing to its importance in realizing the dream of fully self-driving cars and its importance as a key driver-assist feature, especially in highly congested traffic conditions. In this paper, we consider the on-ramp merging problem, in which a single ego vehicle must start on a one-lane entrance ramp and merge with the traffic on a highway (see \cite{wang2018autonomous} and \cite{lin2019anti}).

We first focus on understanding how the two popular classes of motion planning techniques, Model Predictive Control (MPC) and Reinforcement Learning (RL), perform for this problem. In most self-driving systems today, traditional model-based MPC methods are used for motion planning (see \cite{fan2018baidu}). The model can either be learned (see \cite{aswani2013provably}, \cite{kabzan2019learning} and \cite{koller2018learning}) or assumed to be a simple predictor. The reason that model-based MPC methods are widely deployed is because they can guarantee safety, although they are dependent on models and may not be able to plan as effectively over long horizons. On the other hand, RL-based techniques have only recently gained traction in the domain of self-driving, primarily due to their recent astonishing success in various other domains (see \cite{mnih2013playing}, \cite{silver2016mastering} and \cite{silver2017mastering}). Despite being relatively recent, RL-based techniques have been investigated for both motion planning in general (see \cite{aradi2020survey} for a survey) and on-ramp merging in particular (see \cite{wang2018autonomous}, \cite{lin2019anti} and \cite{nishi2019merging}). As we will demonstrate in our experimental results section, since RL agents are trained using copious amounts of data to maximize some notion of long-term reward (which is typically a function of passenger comfort, safety and efficiency, defined later in the paper) and do not just focus on safety, their safety performance is sub-par. Although RL is unable to achieve absolute safety like MPC methods, our simulations indicate that it provides comparatively greater control over optimizing other aspects of the system.  

Although there has been a lot of work combining MPC with reinforcement learning and other learning-based techniques (see \cite{williams2017information}, \cite{zhang2016learning}, \cite{zanon2020safe} and \cite{kamthe2018data}), to the best of our knowledge there is no such work done for the problem of on-ramp lane merging. Our first goal in this paper is to establish whether there exists a trade-off between using RL and MPC methods. To this end, we implement a DDPG-based RL agent (DDPG is a popular RL paradigm and a natural choice due to the continuous action space in our problem; see \cite{lillicrap2015continuous} for more details) and a MPC-based planner (see \cite{fan2018baidu}). To quantify the performance of these two methods, we test them over several thousand episodes/trials of trying to merge on to a highway and use the following metrics to compare them:

\begin{enumerate}
    \item \textbf{Crash Rate} - The fraction of the total episodes in which the ego vehicle crashed while trying to merge on to the highway.
    \item \textbf{Passenger Comfort} - The mean of the absolute instantaneous jerk values of the ego vehicle, averaged over all the episodes. Jerk is the derivative of acceleration and is commonly used as an indicator of passenger comfort (with high jerk values being uncomfortable and low jerk values being comfortable).
    \item \textbf{Efficiency} - The total time the ego vehicle takes to merge on the highway from its starting point, averaged over all the episodes.
\end{enumerate}

After gaining insights into the performance of the DDPG-based RL agent and the MPC-based planner, and establishing a trade-off between them, we aim to blend these two methods in order to design an algorithm that achieves the best of both worlds, i.e., the RL agent's desirable performance in terms of efficiency and passenger comfort, and safety levels close to that of the MPC-based planner. Our main contributions can be summarized as follows:

% \subsection{Motivation and Background}

% \begin{itemize}
%     \item Introduce past solutions and discuss the other RL paper, its advantages and significant limitations. Describe the methodology used, i.e., the specific RL algorithm used. Point out that underlying an RL algorithm is an MDP whose parameters are unknown and the traffic traces are assumed to be sample paths of the Markov chain in the MDP. So, after a lot of training,the algorithm learns to perform well for an unknown MDP which fits the data that it has seen. So inherently, it is not robust to unforeseen traffic patterns. Additionally crashes more than other solutions (why?) 
%     \item RL is a model-free approach, introduce ST solver approach (plus past attempts) and call it a model-based approach which learns a model from sensors and neural networks, assumes the model stays the same in the future, computes a solution, which is implemented for a short time (how many secs?). Then the model is relearned and the process is repeated. Why does it have very low crash rate, but so great on other metrics?
%     \item Talk about the goals with combining them: can we get the best of both worlds?
% \end{itemize}

\begin{enumerate}
    \item Through extensive simulations, we show that DDPG-based RL agents perform better than MPC agents in terms of comfort and efficiency but perform worse than the MPC agents in terms of safety (see Section \ref{ref:expsec}).
    \item Motivated by the above observation, we design a combined RL-MPC agent which performs better than the MPC agent on the comfort and efficiency metrics while matching the MPC agent's safety performance in almost all scenarios. In the few scenarios where the safety of our combined agent is worse than that of MPC alone, the difference is small (see Section \ref{ref:ourappsec}). 
    \item A significant drawback of an RL agent is that it can only learn to perform well in the traffic scenarios on which it is trained. Our combined RL-MPC agent is robust in its ability to deal with traffic scenarios that are different from the training scenarios (see Section \ref{ref:expsec}).
\end{enumerate}

\section{PROBLEM FORMULATION}\label{sec:probformulation}

\begin{figure}
    \centering
    \includegraphics[width=\linewidth]{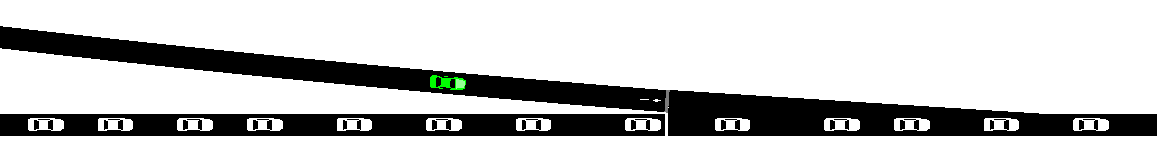}
    \caption{A portion of the road network used by SUMO in our experiments. The ego car is green and the highway traffic is white.}
    \label{fig:sumo_road}
\end{figure}

We consider a ramp merging task consisting of a one-lane entrance ramp merging on to a highway. The ego car, which begins on the entrance ramp about 160 meters from the merge point, must merge with the highway traffic and then travel about 50 meters after merging. A visualization of the highway and on-ramp used for our experiments can be found in Figure \ref{fig:sumo_road}.

The road network and all vehicles are modeled by the Simulation of Urban Mobility (SUMO) simulator \cite{SUMO2018}.  The simulation uses a tick length of 0.2 seconds. At each tick, the next position of each vehicle is calculated using its current speed. If the ego car collides with any other car, the ego car is removed from the network. The ego car is also given a time limit to complete the merging task. If it fails to merge within 100 seconds, the ego car is removed and the current episode ends. The ego car is controlled by our agent, and the other vehicles are allowed to react to the ego car's behavior, braking to avoid collision or accelerating to a desired speed. SUMO offers a variety of driver models to simulate the behavior of the highway traffic. For this paper, we use the Krauss driver model \cite{krauss1998microscopic}, which is the default driver model implemented by SUMO. The algorithm presented in this paper to combine MPC and RL can easily integrate other common driving models, such as the intelligent driver model (IDM). Further, we expect the results of this paper to be largely agnostic to the driver model used, as our primary concern is the comparison and combination of MPC and RL.

%Although Krauss may not be the most common driver model used in simulations, it was convenient for our experiments and We expect the results of this paper to be agnostic to the driver model used, as our primary concern is the comparison and combination of MPC and RL.

We have four main metrics of interest for the on-ramp merging problem: merge rate, crash rate, mean absolute jerk, and time to merge. Merge rate is defined as the probability that the ego car successfully completes the merge. Crash rate is the probability of crashing. For all experiments in this paper, the sum of merge rate and crash rate is $1,$ so we will only report crash rate. The mean absolute jerk is a measure of smoothness of the ego car's trajectory, which we use to measure passenger comfort. Finally, time to merge refers to the mean time required to complete an episode, given that the ego car successfully merges. This is a measure of the efficiency of the ego car's trajectory.

Given these four metrics, our primary task is to control the ego car in such a way that both the mean absolute jerk and time to merge are as small as possible, while ensuring a crash rate close to 0\% and merge rate close to 100\%. In order to perform this task, the ego vehicle is assumed to have perfect knowledge of its position, speed, and acceleration, as well as those of all other cars within 125 meters. In a real autonomous driving system, these values would be estimated using data collected from a variety of sensors installed on the ego vehicle. However, to isolate the planning element of our task, we assume this estimation has been perfectly performed.

\section{MPC AND RL OVERVIEW}

We now give a brief introduction to the MPC and RL approaches we will consider in this paper. MPC and RL-based approaches have both been studied for on-ramp merging and other autonomous driving tasks, although to the best of our knowledge, their relative strengths and weaknesses have not been studied in the context of on-ramp merging. We have adapted these MPC and RL techniques to the on-ramp merging task and the specific implementation details can be found in Section \ref{ref:secimp}.

\subsection{MPC Overview}

\begin{figure}
    \centering
    \includegraphics[width=0.85\linewidth]{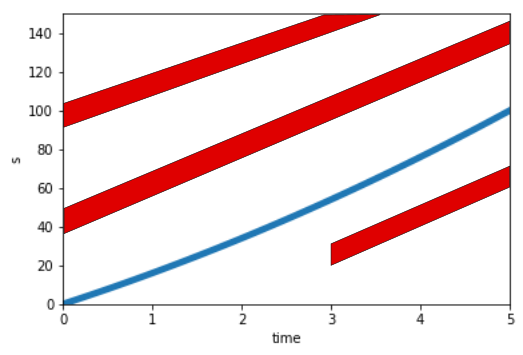}
    \caption{The S-T path planner first uses a (possibly imperfect) model to predict the trajectories of all obstacles in S-T space (red). Then, an optimal trajectory for the ego car is planned to avoid all obstacles (blue). Obstacles only appear on the S-T space once they enter the path of the ego vehicle.}
    \label{fig:st_space}
\end{figure}

For our MPC approach, we chose a dynamic programming-based ST (space-time) solver, borrowing the terminology from \cite{fan2018baidu}. This speed planning algorithm first predicts the trajectories of nearby vehicles over the planning time horizon, then projects those trajectories into the S-T space, as in Figure~\ref{fig:st_space}. In the S-T space, the $t$ axis represents future time, and the $s$ axis represents distance along the path of the ego vehicle, where the ego vehicle starts at $(0,0)$. Thus, the speed planning problem over the next $5$ seconds reduces to finding a collision-free path to the line $t=5$ in S-T space which optimizes some objective function. We use the same objective function as in \cite{fan2018baidu}, incorporating penalties for speed ($s'(t)$), acceleration ($s''(t)$), jerk ($s'''(t)$), and distance to the nearest obstacle ($d_\text{min}(s(t))$):
\begin{align*}
    \min_{s(t)} \int_{0}^{t_{\text{max}}} \left( w_1 \mathbb{I}_{d_\text{min}(s(t)) < d_c} + \frac{w_2}{d_\text{min}(s(t))}\mathbb{I}_{d_\text{min}(s(t)) \geq d_c} \right)dt + \\
    \int_{0}^{t_{\text{max}}}\left( w_3(s'(t) - v^*)^2 + w_4 (s''(t))^2 + w_5 (s'''(t))^2\right)dt
\end{align*}
In the above equation, $w_1, w_2, ..., w_5$ denote the relative penalty weights for getting within an unsafe distance $d_c$ of an obstacle, being far away from an obstacle, differing from the fixed desired velocity $v^*$, and for high acceleration and jerk respectively. The minimization is taken over collision-free trajectories. In order to make the optimization computationally feasible, we discretize the S-T space into a lattice and use dynamic programming to find the optimal trajectory over lattice points.

The prediction of other vehicles' trajectories relies on a model of their behavior. We assume we have access to such a model $\hat{f}$, which can take the current state of the highway $x$ as input, along with the current ego car's action $u$, and return a predicted next state $\hat{f}(x, u)$. To prevent explosive computational complexity in the ST solver, a simple lane following behavior for the ego car is used to make a static prediction of the other cars' trajectories, shown in red in Figure~\ref{fig:st_space}. The model $\hat{f}$ is not assumed to be perfect, and for the purposes of this paper is assumed to be deterministic. Future work could consider more complex MPC approaches.

\subsection{RL Overview}

To compare to and combine with the ST solver, we also implement an RL-based approach for lane merging. Reinforcement learning learns the best way to complete a task by interacting with its environment and observing feedback (see \cite{sutton2018reinforcement} for a more thorough treatment). Specifically, if $x_k \in \mathcal{X}$ is the state of the system at time $k$ and the system evolves noisily as $x_{k+1} = f(x_k, u_k, w_k)$, where $u_k \in \mathcal{U}$ is a control or action and $w_k$ is random noise, then the goal of our RL agent is to find a policy $\pi: \mathcal{X} \to \mathcal{U}$ that maximizes the expected discounted future reward, or value function,
\begin{equation*}
    V_\pi(x) = \mathbb{E}[ \sum_{k=0}^\infty \gamma^k r(x_k, u_k) | x_0 = x],
\end{equation*}
where $u_k = \pi(x_k)$ and $r(x, u)$ is the reward obtained when taking action $u$ in state $x$, and $\gamma \in [0, 1)$ is a discount factor to favor immediate rewards over distant rewards. $V(x) = \sup_{\pi} V_\pi(x)$ is used to denote the optimal value function. Many RL algorithms use the state-action value function, or Q-function $Q_\pi(x, u) = \mathbb{E}[r(x_0, u_0) + \sum_{k=1}^\infty \gamma^k r(x_k, u_k) | x_0 = x, u_0 = u]$ instead of $V_\pi(x)$. The Q-function gives the reward obtained by taking action $u$ at state $x$ and then following policy $u_k = \pi(x_k)$ for $k\geq 1$.

For our lane merging task, we use the following MDP definition (for more details, see section VI):

\begin{itemize}
	\item State representation: the position, velocity, and acceleration of the ego vehicle and all other nearby vehicles;
	\item Action: the jerk of the ego vehicle over the next time step; and
	\item Reward: the following simple reward function, which rewards successful merges but penalizes crashes, incurring jerk, and slow merging:
\end{itemize}
\begin{equation}
	r(x, u) = w_1 s(x, u) - w_2 - w_3 u^2.
\end{equation}

The reward function $r(x, u)$ presented above is similar to reward functions used in other RL approaches for episodic autonomous driving tasks such as in \cite{isele2018navigating}. $w_1, w_2, $ and $w_3$ represent tunable positive weights for success/crashes, efficiency, and jerk, respectively. $s(x, u)$ denotes the success of the episode and takes the value 1 for a successful merge or -1 for a crash. The action $u$ is the constant jerk of the ego car over the next time step. 

Note that this reward function is different from that used by the ST solver. The RL reward function incorporates the success function $s(x, u)$ to provide a sparse, long-term reward for the full episode. Combining this with the remaining components is sufficient for the RL agent to learn both safety and efficiency. However, the ST solver cannot easily incorporate such long-term reward components because it is computationally limited to shorter time horizons. Also, even if the ST solver and RL approaches used the same reward function, they would not find the same optimal solution because of the limited horizon of the ST solver and its static prediction of the behavior of other vehicles. This simpler reward function is convenient for reinforcement learning, as it takes values in a fairly predictable range, penalizes crashes harshly enough for optimal behavior to completely avoid crashes, and still includes all of our metrics of interest. Ultimately, while the RL reward function differs from the ST solver's cost function for the aforementioned reasons, our experiments rigorously ensure that both algorithms were tested on the same safety, comfort, and efficiency criteria.

To learn a good policy $\pi$, we choose the deep deterministic policy gradient (DDPG) algorithm. DDPG is an RL algorithm in a class of algorithms called actor-critic algorithms. It maintains a policy neural network $\pi_{\theta_\pi}(x)$ to represent the agent's policy and a value neural network $Q_{\theta_Q}(x, u)$ to estimate the Q-function for the current policy $\pi_{\theta_\pi}(x)$ ($\theta_\pi$ and $\theta_Q$ are network weights). To train a DDPG agent, training samples of $x_k, u_k, $ and $r(x_k, u_k)$ are collected using a noisy version of the current policy $\pi_{\theta_\pi}$. Those samples are then used to estimate the policy gradient $\nabla_{\theta_\pi} V$ and the Q-function $Q_{\theta_Q}(x, u)$, and the neural network weights are updated through gradient ascent. For more details about DDPG, see \cite{lillicrap2015continuous}.

DDPG is well suited to our on-ramp merging problem due to its ability to handle continuous action spaces and its relative simplicity. Our action space is naturally continuous, as the choice of the ego car's speed can take any real value in a constrained range. Some other popular RL algorithms such as DQN require discrete action spaces, and had worse performance than DDPG in our experiments. DDPG has been widely used to solve autonomous driving tasks (\cite{aradi2020survey, xiong2016combining, wang2018autonomous}), including on-ramp merging. For the on-ramp merging problem, our approach is most similar to that found in \cite{lin2019anti}, as they use a similar highway, DDPG, and also aim to minimize jerk. In contrast, we use a simpler reward function and both compare and combine our DDPG approach with MPC. Other actor-critic RL algorithms such as PPO and A3C are often used for control tasks with continuous action spaces \cite{mnih2016asynchronous, schulman2017proximal}. We did not test these other algorithms in this paper because our focus is to compare RL and MPC and propose a novel algorithm to combine them. We expect similar high-level tradeoffs between MPC and RL for other RL algorithms.

\section{OUR APPROACH}\label{ref:ourappsec}

Through our experiments, we have found that the ST solver and DDPG have complementary advantages and disadvantages, and that combining a trained DDPG agent with an ST solver can produce an agent that outperforms both the ST solver and DDPG in isolation. The differences between ST solver and DDPG are most apparent in situations where the ego car must interact with other vehicles in order to merge smoothly and efficiently. Such situations expose the limitations of the imperfect predictive model used by MPC, while DDPG can learn how to best interact with other drivers through those interactions, without an explicit model for their behavior. As a downside, the DDPG models tend to be unsafe and brittle to scenarios not encountered in training. To combine the two, we use the ST solver as a guide during agent evaluation, judging the safety of the DDPG agent's actions and taking control in the case of unsafe behavior.

\subsection{Evaluating MPC vs RL}

\begin{figure}
    \centering
    \includegraphics[width=\linewidth]{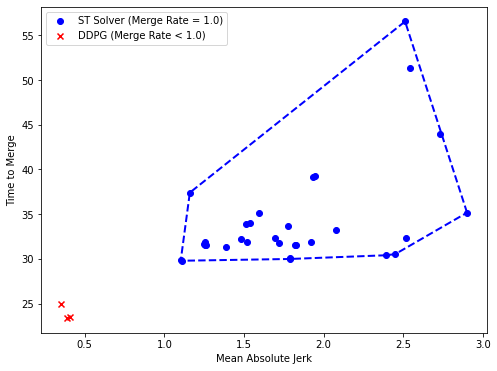}
    \caption{We plot the time to merge vs mean absolute jerk for several DDPG models and ST solver configurations (lower left is better). The ST solver configurations came from a grid search of hundreds of combinations of parameters. The DDPG models we trained lie outside the Pareto boundary of the ST solver's performance and are strictly better in time to merge and mean absolute jerk, but have a nonzero crash rate. ST solver configurations without a 100\% merge rate (usually too conservative to successfully merge) have been omitted.}
    \label{fig:pareto}
\end{figure}

Although our ST solver, like other MPC approaches, claims to find an optimal trajectory, it is only optimal with respect to the predicted trajectories of the highway vehicles. In order to produce an algorithm which can be feasibly solved, these predictions are made independently of the ego vehicle's planned trajectory. Thus, the ST solver cannot easily account for the reactions of the highway traffic to the ego car's behavior. To capture this effect, in our experiments we consider a heavy-traffic merging scenario where the gap between the highway traffic is not large enough for the ego car to safely merge and the highway traffic must slow down to allow the ego car into traffic. Although our simulator SUMO uses the Krauss driver model to predictably control each car, we assume a simpler driver model for our ST solver's predictive model, as in general the behavior of the highway traffic will be unknown and have some error. With these assumptions, we have found that no set of ST solver parameters could achieve the level of efficiency or comfort achieved by our DDPG agents. We tested many ST solver parameters along a grid search to estimate the Pareto optimal front for time to merge and mean absolute jerk, and our trained DDPG agents lay outside this boundary for each of the three seeds we tested, as seen in Figure~\ref{fig:pareto}. For our experiments, we chose an ST solver configuration along the boundary with a good balance of efficiency and comfort.

Although the DDPG agents tend to have better performance than those of the ST solver in terms of efficiency and comfort, they tend to have a nonzero crash rate, and their performance may suffer when evaluated on situations that were not encountered during training. These results, which we experimentally verified (see Section \ref{ref:expsec} for details), are commonly observed for deep RL algorithms. Indeed, many other applications of deep RL to on-ramp merging and other difficult autonomous driving tasks also fail to achieve a crash rate of close to 0\%, despite outperforming other techniques on metrics other than crash rate \cite{bouton2019cooperation, isele2018navigating, bouton2019safe}.

\subsection{Combining MPC with RL}

Because the MPC approach to lane merging can guarantee safety and an RL-based approach can merge more efficiently and more smoothly, but without a guarantee of safety, we propose a combined agent to merge more efficiently than MPC alone, but with the robustness and safety guarantees close to that of MPC. Because the RL agent is assumed to have better performance than the MPC controller, the MPC controller is used as a supervisor to the RL agent, identifying unsafe behavior and taking control when necessary to prevent collisions. Our combined agent uses the unmodified ST solver and DDPG agents as components, rather than modifying the training or parameters of either agent. To the best of our knowledge this is the first application of a supervised RL agent of this type to the on-ramp merging problem.

The basic algorithm (Algorithm 1) can be succinctly described using the following decision-making process at each time step, assuming the starting observed state of the highway is $x_0$ at the current time step. We will then define each quoted term to make the algorithm precise:

\begin{itemize}
	\item Predict the ``planned RL trajectory'' $x_0, x_1, \ldots, x_n$ for the next $n$ time steps.
	\item Evaluate whether the planned RL trajectory is ``unsafe,'' and use the ST solver to control the ego car, if deemed unsafe.
	\item If the planned ST solver trajectory is ``strictly better'' than the planned RL trajectory, use the ST solver to control the ego car.
	\item Otherwise, use the RL agent to control the ego car.
\end{itemize}

\begin{algorithm}
\SetAlgoLined
\LinesNumbered
\DontPrintSemicolon
\SetKwData{ST}{ST Solution}\SetKwData{This}{this}\SetKwData{RL}{RL Solution}
\SetKwFunction{STSolver}{STSolver}\SetKwFunction{ContainsCrash}{IsUnsafe}\SetKwFunction{Min}{min}\SetKwFunction{Jerk}{MeanAbsoluteJerk}\SetKwFunction{DistanceTraveled}{EgoDistanceTraveled}
\SetKwInOut{Input}{input}\SetKwInOut{Output}{output}\SetKwInOut{FunctionInput}{function}
    \Input{The trained RL policy $\pi$}
    \Input{The starting state $x_0$}
    \Input{The length of the RL rollout $n$}
    \Input{A prediction model $\hat{f}$}
    \Input{A minimum safe distance $d_{\text{min}}$}
    \Input{An ST solver \STSolver that outputs a trajectory of length $t_{\text{max}}$}
    \FunctionInput{\ContainsCrash{$d_{\text{min}}$, $x_0, [x_1, \ldots]$} checks whether the ego car is within distance $d_\text{min}$ of other vehicles in  states $x_0, x_1, \ldots$}
    \Output{Whether to use the ST solver or RL policy}
    \tcp{Check if predicted RL trajectory is safe}
 \For{$i = 1$ \KwTo $n$}{
    $u_{i-1} = \pi(x_{i-1})$\;
    $x_i = \hat{f}(x_{i-1}, u_{i-1})$\;
    \If{\ContainsCrash{$d_\text{min}, x_i$}}{
        \KwRet{\ST}\;
    }
 }
 \tcp{Check if terminal state leads to a crash}
 $s_n, s_{n+1}, \ldots, s_{n+t_{\text{max}}}$ = \STSolver{$x_n$}\;
 \If{\ContainsCrash{$d_\text{min}$, $s_n$, $s_{n+1}, \ldots, s_{n+t_{\text{max}}}$}} {
    \KwRet{\ST}\;
 }
 \tcp{Compare efficiency of ST and RL trajectories}
 $s_0, s_1, \ldots, s_{t_{\text{max}}}$ = \STSolver{$x_0$}\;
 k = \Min{$n, t_{\text{max}}$}\;
 $j_{ST}$ = \Jerk{$s_0, s_1, \ldots, s_k$}\;
 $j_{RL}$ = \Jerk{$x_0, x_1, \ldots, x_k$}\;
 $d_{ST}$ = \DistanceTraveled{$s_0, s_k$}\;
 $d_{RL}$ = \DistanceTraveled{$x_0, x_k$}\;
 \If{$d_{RL} = 0$}{
    \KwRet{\ST}\;
 }
 \eIf{$d_{RL} < d_{ST}$ {\bf and} $j_{RL} > j_{ST}$}{
    \KwRet{\ST}\;
 }{
    \KwRet{\RL}\;
 }
 \caption{ST Solver + RL Combination}
\end{algorithm}

One primary component of this algorithm is predicting the RL trajectory. While our ST solver naturally produces a trajectory for the ego car, the RL agent only produces an action $u_0 = \pi(x_0)$ for the given observation. In order to predict a longer trajectory, we use our prediction model $\hat{f}$ from the ST solver to ``roll out'' a longer trajectory. Given the predicted positions of all other cars and our control, we can determine a predicted observation $x_1 = \hat{f}(x_0, u_0)$ of the state of all cars at the next time step. This is fed into the RL agent, producing the expected next action $u_1 = \pi(x_1)$. This process is repeated, producing a sequence of states, which is our planned RL trajectory (lines 1-3 of Algorithm 1).

Next, we need a way to evaluate whether this planned RL trajectory is unsafe. This process has two components. The first and most straightforward is to check for unsafe conditions in the planned trajectory. If the ego car ever comes within a minimum distance $d_{\min}$ of another car in the planned trajectory, then the RL trajectory is deemed unsafe (lines 4-6). The second component is to use the ST solver to check for a feasible trajectory starting at the predicted end state $x_{n}$. If the ST solver cannot find a safe way to proceed from $s_n$, then the planned sequence of RL actions is deemed unsafe and the ST solver's control is used instead (lines 8-11). If the prediction model is perfect, then this process should theoretically avoid all crashes. However, in the presence of imperfect prediction models, tuning the value of $d_{\min}$ can provide a trade off between safety and performance.

The last component of our algorithm is a comparison of the ST and RL trajectories, to ensure that the ST solver is used if the RL trajectory is too inefficient. For both the ST and RL trajectories, the mean absolute jerk and distance traveled are calculated (lines 14-17). If the ST solver travels positive distance while the RL trajectory remains motionless (lines 18-20), or if the ST solver trajectory has strictly less jerk and strictly more distance traveled (lines 21-23), then the ST solver is used instead. In our experiments, we have found that this additional check can greatly improve the ST + RL combination in some scenarios, such as when the supervision of the ST solver pushes the RL agent into an unfamiliar state.

Collectively, all of these components in algorithm 1 constitute a novel combination of MPC and RL that seamlessly balances efficiency and comfort with safety. In the next section, we will experimentally verify that the combination can correct for the unsafe behavior of learned RL policies to produce an agent superior to either one of its constituents. 

\section{EXPERIMENTAL RESULTS}\label{ref:expsec}

For our experiments, we designed a set of five traffic models, designed to require various degrees of interaction between the ego car and highway traffic in order to merge successfully. The first and most difficult traffic pattern consists of a sequence of cars, each traveling a speed of 7 m/s and spaced uniformly at random from 1.2 to 2.0 seconds apart. These cars are controlled by SUMO using the Krauss driver model, and will brake if necessary to avoid collisions, subject to their own limits on acceleration. Every car is $5$ meters in length, so the space between cars is $3.4$ to $9$ meters. Generally, in order to successfully merge, the ego car will need another car to brake at least slightly to let them in. This is more difficult for the ST solver, which cannot account for this behavior in its planning. In order to add additional randomness to the model, the ego car is introduced at a random initial speed between 5 and 25 $m/s$. All cars are 5 meters in length and limited to a speed of 0 to 30 $m/s$ and an acceleration of $-6$ to $4.5$ $m/s^2$. Although SUMO does not implement jerk limits for the vehicles it controls, our ego car is constrained by a jerk limit of $-5$ to $5$ $m/s^3$.

The remaining traffic models vary this default model by reducing the density of traffic, either by increasing the speed of the highway traffic while maintaining the same time gap between cars, or by increasing the time gap between cars while maintaining the speed of the traffic. If we consider the default model to be the ``heavy traffic'' or ``slow'' model, then the models have the following parameters:
\begin{itemize}
	\item \emph{Heavy/Slow traffic}: Cars $1.2$ to $2.0s$ apart, at $7 m/s$
	\item \emph{Medium traffic}: Cars $1.8$ to $2.6s$ apart, at $7 m/s$
	\item \emph{Low traffic}: Cars $2.4$ to $3.2 s$ apart, at $7 m/s$
	\item \emph{Moderate traffic}: Cars $1.2$ to $2.0s$ apart, at $11 m/s$
	\item \emph{Fast traffic}: Cars $1.2$ to $2.0s$ apart, at $15 m/s$
\end{itemize}

For each traffic model, we trained three RL agents with identical hyperparameters. In this paper, we present the RL agent with the best performance on each traffic model. We used the same ST solver parameters for each model. To choose this ST solver configuration, we performed a grid search to test hundreds of ST solver configurations, then chose a configuration with both 100\% merge rate and the best balanced performance in mean absolute and time to merge.

\subsection{Combination Performance}

\begin{table}
    \caption{Comparing agent performance on various traffic densities}
    \label{tab:traffic_density}
\begin{center}
    \begin{tabular}{|c c c c|}
        \hline
         Metric & RL & ST Solver & Combined \\
         \hline
        \hline
         \multicolumn{4}{|c|}{Heavy Traffic}\\
         \hline
         Crash Rate & 0.225\% & \textbf{0\%} & \textbf{0\%} \\
         Mean Absolute Jerk & \textbf{0.352} & 1.105 & 0.780 \\
         Time to Merge & \textbf{25.00} & 29.84 & 28.79 \\
         \hline
         \hline
         \multicolumn{4}{|c|}{Medium Traffic}\\
         \hline
         Crash Rate & 0.8\% & \textbf{0\%} & \textbf{0\%} \\
         Mean Absolute Jerk & \textbf{0.428} & 1.262 & 0.686 \\
         Time to Merge & \textbf{22.13} & 28.64 & 26.56 \\
         \hline
         \hline
         \multicolumn{4}{|c|}{Low Traffic}\\
         \hline
         Crash Rate & 0.325\% & \textbf{0\%} & \textbf{0\%} \\
         Mean Absolute Jerk & \textbf{0.288} & 1.074 & 0.535 \\
         Time to Merge & \textbf{22.05} & 25.66 & 23.70 \\
         \hline
         \multicolumn{4}{@{}p{0.4\textwidth}@{}}{\footnotesize We compare the crash rate, mean absolute jerk, and time to merge of RL, the ST solver, and our combination of the two on three traffic scenarios, with differing densities of highway traffic. The best algorithm for each metric is bolded. Note that in these scenarios, the combination of ST solver + RL is able to achieve the safety of the ST solver while outperforming the ST solver in comfort and efficiency.}
    \end{tabular}
\end{center}
\end{table}

For each of the above traffic models, we evaluated the performance of the RL agent alone, the ST solver alone, and the combination of RL and ST solver using $d_\text{min} = 5.1$ as the combination's safety parameter (given the length of each car, this threshold ensures that the ST solver will take over if the RL agent expects to come within 0.1 meters of another vehicle). For each of these trials, the merge rate and crash rate summed to 1, so we omit the merge rate. Each agent or combination was evaluated for 4000 episodes. Table \ref{tab:traffic_density} shows the comparison for each agent on the heavy, medium, and low traffic models. Note that as the traffic density increases and merging becomes easier without requiring other cars to slow for the ego car, the performance of the ST solver approaches that of the RL agent alone. Although the RL agents are unable to achieve a crash rate below 0.2\% over 4000 trials, both the ST solver and combined ST + RL agents never experience a crash. The performance of the combined agent is consistently strictly better than the ST solver alone in our metrics of interest.

We also compare each agent on its performance as we vary the speed of the traffic. Increasing the speed also increases the gap between highway cars, making it easier for the ST solver with comfort and efficiency close to that of the RL agent. The results of these experiments can be found in Table \ref{tab:traffic_speed}. Again, the combined ST + RL agents consistently outperform the RL agents alone, though the gap in time to merge becomes negligible in the fastest traffic scenario.

\begin{table}
    \caption{Comparing agent performance on various traffic speeds}
    \label{tab:traffic_speed}
\begin{center}
    \begin{tabular}{|c c c c|}
        \hline
         Metric & RL & ST Solver & Combined \\
         \hline
         \multicolumn{4}{|c|}{Slow Traffic}\\
         \hline
         \hline
         Crash Rate & 0.225\% & \textbf{0\%} & \textbf{0\%} \\
         Mean Absolute Jerk & \textbf{0.352} & 1.105 & 0.780 \\
         Time to Merge & \textbf{25.00} & 29.84 & 28.79 \\
         \hline
         \multicolumn{4}{|c|}{Moderate Traffic}\\
         \hline
         \hline
         Crash Rate & 0.55\% & \textbf{0\%} & \textbf{0\%} \\
         Mean Absolute Jerk & \textbf{0.325} & 1.280 & 0.561 \\
         Time to Merge & \textbf{17.43} & 20.34 & 18.09 \\
         \hline
         \multicolumn{4}{|c|}{Fast Traffic}\\
         \hline
         \hline
         Crash Rate & 0.225\% & \textbf{0\%} & \textbf{0\%} \\
         Mean Absolute Jerk & \textbf{0.431} & 1.153 & 0.737 \\
         Time to Merge & \textbf{13.72} & 14.48 & 14.43 \\
         \hline
    \end{tabular}
\end{center}
\end{table}

\subsection{Out-of-Distribution RL Evaluation}

Since RL agents learn from examples presented to them during training, it is important to evaluate their performance in traffic scenarios that were not present in the training set. While one would hope that the learned policy could generalize to these new situations, the actual performance may be poor in practice. To test our RL agents' abilities to generalize, along with the ability of RL-MPC combination to prevent crashes, we test the agents trained on our moderate and medium traffic distributions on traffic models with different traffic speeds and time between cars, respectively. These results for the agents featured in Tables \ref{tab:traffic_density} and \ref{tab:traffic_speed} are shown in Table \ref{tab:out-of-distribution}.

We find that the performance of the RL agents deteriorates significantly when evaluated on new traffic models. On the other hand, our combined RL-MPC agent has very little loss in performance and is fairly robust to traffic patterns that were not encountered during training. We do observe that our combined RL-MPC agent fails to achieve a 0\% crash rate in some out-of-distribution traffic scenarios, although the crash rate is still quite low in these scenarios. One way to further reduce the crash rate would be to increase the combination's safety parameter $d_\text{min}$, allowing the ST solver to take control sooner in case of dangerous situations. But another approach, and a possible direction for future work, would be to design a controller which senses the traffic distribution and switches to a crash-free ST solver in case the combined RL-MPC agent can not guarantee absolute safety for the sensed traffic distribution.  
\begin{table}
    \centering
    \caption{Evaluating agents trained on different traffic models}
    \label{tab:out-of-distribution}
    \begin{tabular}{|c c c c|}
        \hline
        Metric & RL & ST Solver & Combined \\
        \hline
        \hline
        \multicolumn{4}{|c|}{RL trained on medium traffic, evaluated on heavy traffic}\\
        \hline
        Crash Rate & 21.20\% & \textbf{0\%} & 0.175\% \\
        Mean Absolute Jerk & \textbf{0.479} & 1.105 & 1.128 \\
        Time to Merge & \textbf{23.29} & 29.84 & 24.70 \\
        \hline
        \hline
        \multicolumn{4}{|c|}{RL trained on medium traffic, evaluated on low traffic}\\
        \hline
        Crash Rate & 1.025\% & \textbf{0\%} & 0.05\% \\
        Mean Absolute Jerk & \textbf{0.545} & 1.074 & 1.047 \\
        Time to Merge & \textbf{21.43} & 25.66 & 22.80 \\
        \hline
        \hline
        \multicolumn{4}{|c|}{RL trained on moderate traffic, evaluated on slow traffic}\\
        \hline
        Crash Rate & 85.1\% & \textbf{0\%} & \textbf{0\%} \\
        Mean Absolute Jerk & \textbf{0.838} & 1.105 & 1.196 \\
        Time to Merge & \textbf{23.81} & 29.84 & 24.40 \\
        \hline
        \hline
        \multicolumn{4}{|c|}{RL trained on moderate traffic, evaluated on fast traffic}\\
        \hline
        Crash Rate & 18.8\% & \textbf{0\%} & \textbf{0\%} \\
        Mean Absolute Jerk & \textbf{0.558} & 1.153 & 0.651 \\
        Time to Merge & 15.29 & 14.48 & 15.48 \\
        \hline
        \multicolumn{4}{@{}p{0.4\textwidth}@{}}{\footnotesize As in the previous tables, we compare the crash rate, mean absolute jerk, and time to merge of RL, the ST solver, and our combination of the two on several traffic scenarios, with differing densities of highway traffic. This time, the RL policies are evaluated on a traffic distribution that is different from the one used for training. Alone, these RL policies have a high crash rate. However, combining the RL policy with the ST solver dramatically reduces the crash rate.}
    \end{tabular}

\end{table}

\section{IMPLEMENTATION DETAILS OF MPC AND RL}\label{ref:secimp}

Our ST solver and DDPG implementations were written in Python, using Cython to speed up performance-critical ST solver code. The source code for these implementations and all parameters used for our experiments can be found on GitHub at https://github.com/jlubars/RL-MPC-LaneMerging. The ST solver was implemented ourselves, as we failed to find an existing implementation which we could easily apply to the SUMO simulator. For the DDPG, we used the implementation from \cite{nota2020autonomous}, with its default neural network architecture: the Q-network and policy network are entirely separate and feature a fully connected linear layer with 400 outputs, then a ReLU activation, another fully connected layer with 300 outputs, another ReLU activation, and finally a last fully connected linear layer with 1 output.

For our ST solver, we used the cost function specified in section 3, but with discrete approximations for all derivatives, and using a search over a lattice for the trajectory $s(t)$. The resolution of the lattice is $0.3$ seconds in time and $0.05$ meters in distance, and planning was done over a horizon of 5 seconds and 150 meters. We set $w_1=10,000,000$, $w_2=10$, $w_3=0.5$, $w_4=10$, $w_5=10$, $v^*=30$, and $d_\text{min} = 5$. The closest feasible trajectory with a resolution of the simulator's step size of 0.2 meters was then found using quadratic programming.

For the DDPG, we used the car's jerk as our action, as it retains a consistent range throughout the merging task. The jerk was clipped to the range allowed by our car's speed and acceleration limits. For the reward function, we used a simple slotted reward function inspired by other related work such as \cite{isele2018navigating} while providing incentives for smoothness and efficiency:
\begin{equation*}
r(x, u) = w_1 s(x, u) - w_2 - w_3 u^2
\end{equation*}
Here, $s(x,u)$ represents the success of the merge, taking the value $1$ for a successful merge and $-1$ for a crash. A constant penalty is applied each time step with weight $w_2$, to incentivize a lower time to merge. Finally, because the action $u$ is the current jerk, the third term penalizes high amounts of jerk, to ensure a smoother trajectory. After some experimentation, we chose $w_1 = 10$, $w_2 = 0.02$, and $w_3 = 0.02$, in an attempt to balance the different reward terms while still making it optimal to successfully merge.

Finally, we describe our DDPG's state (or observation) space. Because the road structure is fixed, we do not include information about the road structure in the observation vector. This observation vector contains 20 scalar components. The first four are the ego car's kinematics, consisting of the $x$ and $y$ coordinates of the ego car, along with its scalar speed and acceleration. The remaining 16 components encode the observations of other nearby cars. The road network is oriented such that the highway vehicles travel parallel to the $x$ axis. We consider a highway car to be ahead of the ego car if it has a larger $x$ coordinate than the ego car. Otherwise, the highway car is behind the ego car. If present, the two closest cars ahead and the two closest cars behind the ego car (within the sensor range of 125 meters) each have the following four components in the observation vector:
\begin{enumerate}
	\item Difference in $x$ coordinate to the ego car
	\item Difference in speed to the ego car
	\item Acceleration
	\item 1 if the car is present, otherwise 0 (in which case the other components are also set to 0).
\end{enumerate}
Some other work prefers to leave the acceleration of non-ego cars from the observation vector, assuming that it cannot be estimated accurately. We have found that the DDPG still learns to merge acceptably with this choice of ego vector.

\section{CONCLUSIONS}
We considered the problem of autonomous on-ramp merging and designed an algorithm that outperforms the state-of-the-art in terms of efficiency, comfort, robustness, and safety metrics. Our algorithm is based on two key conclusions that we reached about the state-of-the-art: existing MPC-based algorithms perform well from a safety and robustness perspective, while RL-based algorithms perform well in terms of efficiency and comfort but crash often enough to be unusable in practice. Our main contribution is to combine the algorithms in an effective manner which allows the blended algorithm to perform well on all metrics. The use of MPC to improve the safety of an RL algorithm is a particular example of a safe-RL algorithm. One can consider other alternatives (e.g., \cite{garcia2015comprehensive}) to further improve the performance of our algorithm and this could be a topic for further research. 

Thus far, we have presented results for a single lane merging task, combining a specific MPC approach (dynamic programming-based ST solver) with one RL algorithm (DDPG). Future work could consider more complicated autonomous driving tasks, such as merging on a highway with multiple lanes or using more complex or realistic driver models such as IDM (\cite{treiber2000congested}). Another interesting extension would be to add noisy observations, as measurements of the speed and velocity of other vehicles will not be perfect in practice. While DDPG is one natural choice for our continuous state-space model, there are other algorithms in the literature such as PPO and A3C (\cite{mnih2016asynchronous, schulman2017proximal}) which could also be used. Our MPC approach uses a single model for non-ego cars. More sophisticated MPC approaches could further use an additional inference procedure to select a model for a car's behavior from a collection of available models. Combining other RL and MPC approaches is a topic for further research.

\vskip 1em
\noindent\textbf{Acknowledgments:} Part of this work was done when Harsh Gupta and Joseph Lubars were summer interns at Xsense.ai and Xmotors.ai. The research was also supported in part by the NSF/USDA Grant AG 2018-67007-28379.

\addtolength{\textheight}{-9cm}   % This command serves to balance the column lengths
%\newpage
%%%%%%%%%%%%%%%%%%%%%%%%%%%%%%%%%%%%%%%%%%%%%%%%%%%%%%%%%%%%%%%%%%%%%%%%%%%%%%%%

\bibliography{refs}
\bibliographystyle{ieeetran}

\end{document}